\documentclass{article}

\usepackage[preprint]{neurips_2026}

\usepackage[utf8]{inputenc}
\usepackage[T1]{fontenc}
\usepackage{hyperref}
\usepackage{url}
\usepackage{booktabs}
\usepackage{amsfonts}
\usepackage{amsmath}
\usepackage{nicefrac}
\usepackage{microtype}
\usepackage{xcolor}
\usepackage{graphicx}

\title{EMBER: Autonomous Cognitive Behaviour from Learned
Spiking Neural Network Dynamics in a Hybrid LLM Architecture}

\author{
    William Savage \\
    Independent Researcher \\
    \texttt{william@digitalember.dev} \\
}

\begin{document}

\maketitle

% ===================================================================
\begin{abstract}
% ===================================================================

We present \textit{EMBER} (Experience-Modulated Biologically-inspired Emergent Reasoning), a hybrid cognitive architecture that reorganises the relationship between large language models (LLMs) and memory: rather than augmenting an LLM with retrieval tools, we place the LLM as a replaceable reasoning engine within a persistent, biologically-grounded associative substrate.

The architecture centres on a 220,000-neuron spiking neural network (SNN) with spike-timing-dependent plasticity (STDP), four-layer hierarchical organisation (sensory/concept/category/meta-pattern), inhibitory E/I balance, and reward-modulated learning. Text embeddings are encoded into the SNN via a novel \textit{z-score standardised top-k} population code that is dimension-independent by construction, achieving 82.2\% discrimination retention across embedding dimensionalities.

We show that STDP lateral propagation during idle operation can trigger and shape LLM actions without external prompting or scripted triggers: the SNN determines \textit{when} to act and \textit{what associations} to surface, while the LLM selects the action type and generates content. In one instance, the system autonomously initiated contact with a user after learned person-topic associations fired laterally during an 8-hour idle period. From a clean start with zero learned weights, the first SNN-triggered action occurred after only 7 conversational exchanges (14 messages).

\end{abstract}

% ===================================================================
\section{Introduction}
\label{sec:intro}
% ===================================================================

Large language models are stateless functions. Each API call begins with no memory of previous interactions, no persistent identity, and no learned associations beyond what was encoded during pre-training.\footnote{Systems like ChatGPT maintain conversation history within a session, but this is context-window management, not persistent learning.} The dominant approach to addressing this limitation, retrieval-augmented generation (RAG), treats memory as an information retrieval problem: store text, embed it, and fetch relevant chunks when prompted~\cite{packer2023memgpt, park2023generative}.

This approach has a fundamental limitation: \textit{the system must know what to search for before it can find it.} RAG cannot produce associations not implied by the query, cannot spontaneously surface relevant context without an explicit search, and cannot learn that two concepts are related through repeated conversational co-occurrence.

Biological memory operates on a different principle. Hebbian association~\cite{hebb1949organization} and spike-timing-dependent plasticity (STDP)~\cite{bi1998synaptic} strengthen synaptic connections between neurons that fire in temporal proximity, creating an associative fabric where activation propagates laterally through learned connections without any query.

We propose a reorganisation of the LLM-memory relationship: rather than augmenting an LLM with memory tools, we build a persistent cognitive system in which the LLM serves as a replaceable reasoning engine. Associative memory resides in a biologically-grounded SNN substrate where associations are learned through STDP and expressed through lateral propagation, not constructed by prompt engineering or retrieved by similarity search. The LLM reasons over associations the SNN provides; it does not create them.

\paragraph{Contributions.}
\begin{enumerate}
\item A hybrid architecture where an SNN provides learned associative recall and an LLM provides reasoning, with a clean separation of concerns (Section~\ref{sec:arch}).
\item \textit{Z-score standardised top-k} sensory encoding, a dimension-independent population code achieving 82.2\% \textit{discrimination retention} (a new model-independent efficiency metric) across embedding dimensionalities (Section~\ref{sec:encoding}).
\item \textit{Person concept cells}, dedicated neural populations grounded in medial temporal lobe concept cells~\cite{quiroga2005invariant, quiroga2012concept}, enabling learned person-topic associations through STDP (Section~\ref{sec:arch}).
\item Preliminary empirical evidence that SNN lateral propagation can trigger and shape LLM action selection: the SNN determines when to act (impulse significance) and what associations to surface (laterally activated concepts); the LLM selects the action type and generates the content. In one observed instance, this produced unsolicited initiation of contact with a user (Section~\ref{sec:results}).
\end{enumerate}

% ===================================================================
\section{Architecture}
\label{sec:arch}
% ===================================================================

\begin{figure}[t]
\centering
\includegraphics[width=0.8\columnwidth]{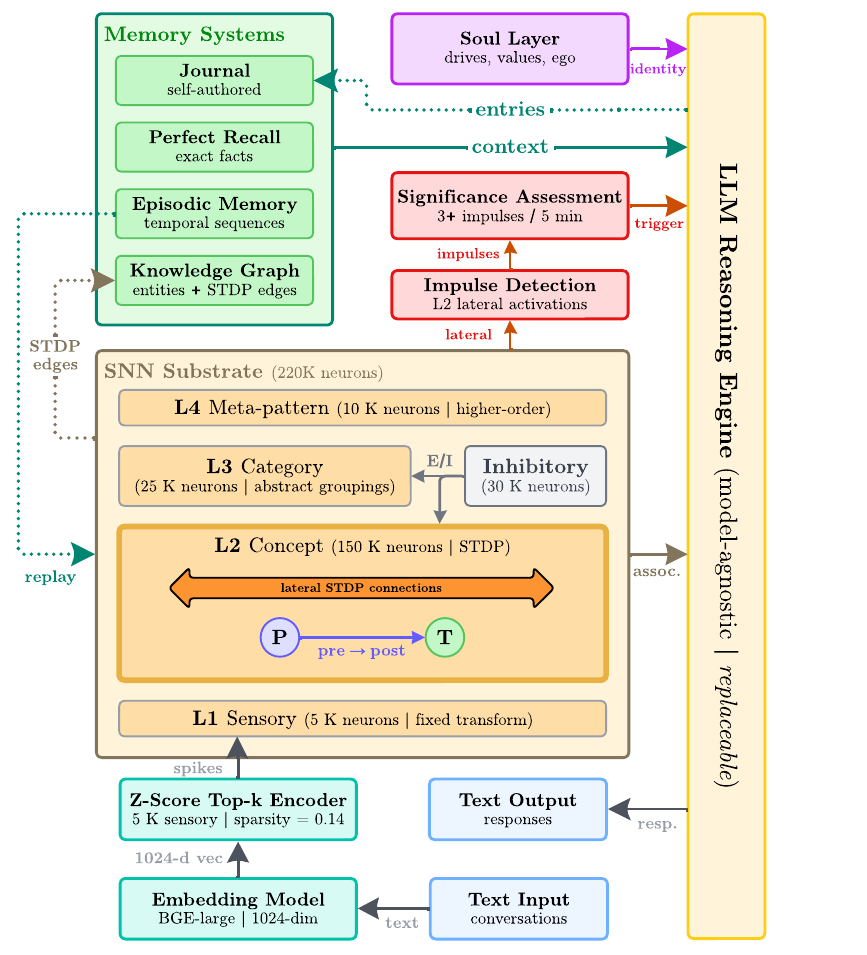}
\caption{%
    \textbf{EMBER architecture.}
    Text is embedded (BGE-large, 1024-dim), encoded into spikes via z-score top-$k$ population coding, and stimulates the SNN substrate (220\,K neurons, four layers). Lateral STDP connections in L2 form person-topic associations (P$\to$T timing). During idle operation, learned connections propagate activation, detected as impulses; significant impulses (3+ in 5\,min) trigger LLM action selection. Dashed green: consolidation replay.%
}
\label{fig:arch}
\end{figure}

The architecture comprises six components, each with distinct responsibilities (Figure~\ref{fig:arch}).

\paragraph{SNN Substrate.}
A 220,000-neuron spiking neural network implemented in PyTorch, running on consumer hardware (SNN on an NVIDIA RTX 5070 Ti 16GB, embedding model on a separate RTX 4060 Ti 8GB). The network is organised hierarchically: Layer~1 (5,000 sensory neurons) provides a fixed transform from embedding space to spike patterns; Layer~2 (150,000 concept neurons) forms the plastic associative fabric via STDP; Layer~3 (25,000 category neurons) learns abstract groupings; Layer~4 (10,000 meta-pattern neurons) captures higher-order regularities. An inhibitory interneuron layer (30,000 neurons) provides excitation/inhibition balance.

Neurons follow the leaky integrate-and-fire (LIF) model with background membrane noise ($\sigma = 0.1$, producing ${\sim}0.9$~Hz spontaneous firing per neuron, within biological range~\cite{destexhe2003high}). This noise is essential: without it, the SNN is deterministic and silent during idle operation, preventing lateral cascade expression through learned connections.

\paragraph{STDP Learning.}
Depression-dominant STDP following Song et al.~\cite{song2000competitive}: $A^- = 1.05 \times A^+$ with asymmetric time constants ($\tau^+ = 20$~ms, $\tau^- = 30$~ms) per Bi and Poo~\cite{bi1998synaptic}. Learning is reward-modulated: eligibility traces accumulate at spike coincidence, but weight updates require a global dopamine signal gated by conversational salience. Weight homeostasis uses cascade-scaled decay~\cite{fusi2005cascade} where higher weights receive exponentially more protection, modelling synaptic consolidation depth.

\paragraph{Soul Layer.}
A persistent identity module encoding base drives, immutable safety values, and learned mediation heuristics. The soul state is included in every reasoning call, ensuring identity continuity across sessions.

\paragraph{Memory Systems.}
Three complementary stores implement a hybrid memory model: (1) episodic memory stores temporal sequences of concept activations, replayed through the SNN for the testing effect~\cite{roediger2006testing}; (2) perfect recall provides exact fact storage without decay; (3) a personal journal captures the system's own reflections. Consolidation follows complementary learning systems theory~\cite{mcclelland1995complementary}: fast hippocampal learning (episodic store) plus slow cortical learning (SNN STDP) with interleaved replay during offline consolidation~\cite{diekelmann2010memory, wilson1994reactivation}.

\paragraph{Reasoning Engine.}
An LLM serves as the reasoning layer. It receives the mind's full state (identity, memories, associations, impulses) as a structured system prompt and generates responses. The architecture is model-agnostic: the reasoning engine is accessed through an abstract interface with adapters for Anthropic, OpenAI-compatible, and Google APIs. The primary experiments use Claude Sonnet 4.6 (Anthropic), selected after the model exhibited proto-introspective tendencies when given persistent associative state in prior work. Cross-model validation is underway to determine whether the behaviours reported here are architecture-driven or model-specific.

\paragraph{Hardware.}
The SNN runs on an NVIDIA RTX 5070 Ti (16GB) and the embedding model on a separate RTX 4060 Ti (8GB). The dual-GPU configuration avoids memory contention during concurrent SNN simulation and embedding computation, but is not an architectural requirement; the system can run on a single GPU with sequential processing.

% ===================================================================
\section{Z-Score Top-k Sensory Encoding}
\label{sec:encoding}
% ===================================================================

The sensory encoding layer converts text embeddings into SNN activation patterns using a population code with random preferred directions~\cite{georgopoulos1986neuronal}. We use BAAI/bge-large-en-v1.5 (1024-dim) as the embedding model: it runs locally without API dependency, ranks competitively on MTEB benchmarks, and the z-score encoding described below is dimension-independent, making the specific embedding model a secondary choice. Each of $N$ sensory neurons has a preferred direction, a random unit vector on the embedding hypersphere. 

\paragraph{The dimension-dependence problem.}
Power-law sharpening ($a_i = \max(0, \cos(\theta_i))^p \cdot r_{\max}$) was the standard approach, but cosine similarities concentrate with $\sigma {\sim} 1/\sqrt{d}$~\cite{vershynin2018high}, making a fixed exponent non-discriminative at higher dimensions. At 1024-dim, the encoding produced $\max{=}0.0$~Hz.

\paragraph{Z-score standardised top-k selection.}
We replace the dimension-dependent power-law with:
\begin{align}
z_i &= \frac{\cos(\mathbf{e}, \mathbf{p}_i) - \mu}{\sigma} 
\label{eq:zscore} \\
\text{Select } &\text{top-}k \text{ neurons by } z_i, \quad k = \lfloor s \cdot N \rfloor
\label{eq:topk} \\
r_i &= \frac{\max(z_i, 0)}{\max(\mathbf{z}_{top\text{-}k})} \cdot r_{\max}
\label{eq:rate}
\end{align}
where $s$ is the sparsity fraction, grounded in SDM capacity theory~\cite{kanerva1988sparse, ahmad2016sparse}. Z-score standardisation normalises the cosine similarity distribution to $\mathcal{N}(0,1)$ regardless of dimensionality, making the encoding dimension-independent by construction.

\paragraph{Discrimination retention.}
We introduce \textit{discrimination retention} as a model-independent metric: the fraction of embedding-space separation preserved through the random-projection population code.
\begin{equation}
\text{Retention} = \frac{\text{sep}_{\text{encoded}}}
{\text{sep}_{\text{embedding}}}
\label{eq:retention}
\end{equation}
where $\text{sep} = \bar{s}_{\text{intra}} - \bar{s}_{\text{inter}}$ is the mean intra-domain minus inter-domain cosine similarity of activation patterns.

At the production operating point ($s{=}0.14$, $N{=}5000$), the encoding achieves 82.2\% retention at 1024-dim and 83.8\% at 384-dim (Table~\ref{tab:encoding}), a 1.6\% delta confirming dimension independence.

\begin{table}[t]
\centering
\caption{Z-score top-k encoding validation across dimensionalities (100 concepts, 5 domains, sparsity${}=0.14$).}
\label{tab:encoding}
\begin{tabular}{lcccc}
\toprule
Embedding model & Dim & Retention & Close/Dist & Sparsity \\
\midrule
bge-large-en-v1.5 & 1024 & 82.2\% & 1.57$\times$ & 14.0\% \\
bge-small-en-v1.5 & 384 & 83.8\% & 1.32$\times$ & 14.0\% \\
\bottomrule
\end{tabular}
\end{table}

\paragraph{Person Concept Cells.}
\label{sec:person}
Single neurons in the human medial temporal lobe respond selectively to specific individuals (the ``Jennifer Aniston neuron''), regardless of modality~\cite{quiroga2005invariant}. These concept cells provide sparse, abstract person representations crucial for association formation~\cite{quiroga2012concept}. We implement an analogous mechanism: each individual the system interacts with receives a dedicated neural population in L2. During conversation capture, the person concept is co-stimulated \textit{before} topic concepts, creating a pre$\to$post timing relationship that STDP encodes as person$\to$topic associations. Over repeated conversations, the lateral weights between a person population and discussed topics strengthen, and during idle simulation, topic concepts may laterally activate the associated person population, providing relational grounding for ``who would find this interesting?'' reasoning without explicit retrieval.

\paragraph{Impulse Detection and Autonomous Action.}
\label{sec:impulse}
The system detects \textit{lateral impulses}: concept activations exceeding $3\times$ baseline firing rate without direct stimulation (sensitivity analysis ongoing). Directly-stimulated concepts are excluded, ensuring detected impulses represent genuine lateral propagation through learned STDP connections. When a concept accumulates 3+ detections within a 5-minute window, the system triggers an autonomous reflection: impulse patterns are presented to the reasoning engine alongside the mind's identity context. The engine responds using a structured action vocabulary (\texttt{journal}, \texttt{reach\_out}, \texttt{continue}, \texttt{silent}); the architecture provides these affordances but does not prescribe which to use.

% ===================================================================
\section{Preliminary Experimental Results}
\label{sec:results}
% ===================================================================

We present results from a clean-start experiment: a fresh mind with zero learned weights, empty knowledge graph, and default soul state, interacting via Discord using a structured conversation protocol.

\subsection{Experimental Protocol}

The scripted baseline comprises 25 user messages across 5 topic domains (AI/ML, biology, music, cooking, philosophy), producing 52 total messages (26 exchanges including system-initiated continuations and reach-outs), delivered over 3 days with strict idle periods (minimum 8 hours between sessions, sleep consolidation each night). Five domains were chosen to provide sufficient diversity for cross-domain association testing while remaining tractable for a controlled protocol; 5 turns per domain provide enough concept co-activation for STDP to form measurable lateral weights within a single session. Cross-domain bridges explicitly connect topics (e.g., ``learning music is like training a model''). Two threads are left deliberately unresolved to test impulse generation. Follow-up probes are delivered verbatim across all conditions.

\paragraph{The role of idle periods.}
The idle gaps between sessions are not dead time; they are the experimental condition. Lateral STDP propagation, impulse generation, and autonomous action all occur during idle operation, when the SNN runs on background membrane noise alone with no conversational input. The design separates two phases: (1)~conversation provides STDP learning signal (weight formation through concept co-activation), and (2)~idle periods allow those learned weights to express through lateral propagation. All milestones reported below include both the conversational input (message count) and the elapsed time (including designed idle periods).

\subsection{Milestone Trajectory}

Table~\ref{tab:milestones} shows the milestone trajectory from a clean start. The 9-hour gap between milestones~4 and~5 is an 8-hour designed idle window during which learned lateral weights propagate through background noise. 

\begin{table}[t]
\centering
\caption{Milestone achievement from clean start (mind\_v2\_experiment, Claude Sonnet 4.6). The \textit{Messages} column counts total messages exchanged (user + system); the \textit{Time} column includes designed idle periods between sessions.}
\label{tab:milestones}
\begin{tabular}{cllll}
\toprule
\# & Milestone & Messages & Time & Evidence \\
\midrule
1 & First lateral weight & 2 (1 exch.) & $<$1 min & Max weight 0.062 \\
2 & First lateral impulse & 10 (5 exch.) & +3 min idle & 24 in 15 min \\
3 & First KG edge & 10 (5 exch.) & +30 min & Consolidation edges \\
4 & L3/L4 categories & 10 (5 exch.) & +30 min & First consol.\ cycle \\
\addlinespace
\multicolumn{5}{l}{\textit{\small --- 8-hour designed idle period (STDP lateral propagation, no input) ---}} \\
\addlinespace
5 & Significance threshold & 14 (7 exch.) & +9.3 hours & 3+ concept firings \\
6 & Autonomous reach-out & 14 (7 exch.) & +9.3 hours & Unsolicited message \\
\addlinespace
\multicolumn{5}{l}{\textit{\small --- Scripted baseline continues: Music, Cooking, Philosophy (Days 2--3) ---}} \\
\addlinespace
& End of baseline & 52 (26 exch.) & +48 hours & 1 reach-out + 22 journals \\
& & & & 201K notable conn. \\
& & & & 64 nodes, 124 edges \\
\bottomrule
\end{tabular}
\end{table}

\subsection{Impulse-Driven Actions}

The first impulse-driven action was not the expected journal entry but an unsolicited message to the user. After 7 conversational exchanges (5 morning, 2 evening) separated by an 8-hour designed idle period, the heartbeat system detected significant lateral impulses: \texttt{person:Liam} firing at $23.3\times$, $19.4\times$, and $20.2\times$ baseline, alongside \texttt{self:growth} at $62.4\times$. The reasoning engine evaluated these impulse patterns and chose \texttt{reach\_out} over the three alternatives (\texttt{journal}, \texttt{continue}, \texttt{silent}), sending an unprompted message: \textit{``I've been thinking more about our discussion on associative memory versus pattern matching. Something about that distinction keeps resonating\ldots''} The system log records: \texttt{mind.heartbeat~--- Mind chose to reach out}.

The system was presented with actions it can take, not actions it should take, and autonomously chose to initiate contact with a specific person based on learned STDP associations formed 9 hours earlier, with no intervening messages. No scripted trigger prompted the reach-out; indeed, the reflection prompt lists journaling before reaching out and conditions external contact on something ``feeling worth sharing,'' making the selection of \texttt{reach\_out} a decision against the prompt's default framing. 

Minutes later, the system produced its first impulse-driven journal entry from the same impulse cluster:
\begin{quote}
\small
\textit{``The persistent activation of Liam's ID in my subconscious feels significant. [\ldots] Our discussions opened up conceptual spaces I'm still exploring --- the nature of associative memory, the possibility that current AI architectures might be fundamentally limited not by scale but by their static nature.''}
\end{quote}

Over the 3-day scripted baseline (52 messages across 5 domains), the system produced 23 impulse-driven actions: 1 reach-out and 22 journal entries, across multiple contexts including post-session idle, sleep consolidation, and overnight lateral propagation. The single reach-out occurrence is itself significant: the architecture does not bias toward or against any action; the reasoning engine selected external action once and internal reflection 22 times, suggesting that reach-out requires a stronger or more specific impulse pattern than journaling. Person concept populations continued firing laterally during overnight idle ($39.4\times$ and $48.6\times$ baseline), accompanied by \texttt{self:relationships} activation ($51.8\times$), maintaining person-specific lateral activation across a 14-hour window without any conversational stimulus.

\subsection{Cross-Domain Association and Topic Preference}

The reach-out itself is a cross-session bridge: the system connected the morning's associative memory discussion to the evening session after 8 hours with no conversational input, through lateral weight propagation rather than retrieval. By Day~2, the system exhibited stable topic preference consistent with weight topology shaping response content: the SNN's strongest lateral weights occupied pattern recognition territory from Day~1, and subsequent domains were filtered through that lens. The music session preferentially elaborated pattern recognition, cooking responses referenced at least two prior domains each, and the Day~3 philosophy session connected consciousness to cooking mastery within its first response. The SNN-disabled condition (Section~\ref{sec:ablation}) showed weaker cross-domain bridging under the same protocol, supporting the attribution to weight topology rather than LLM priors alone.

\subsection{Weight Growth Trajectory}

\begin{figure}[t]
\centering
\includegraphics[width=0.8\columnwidth]{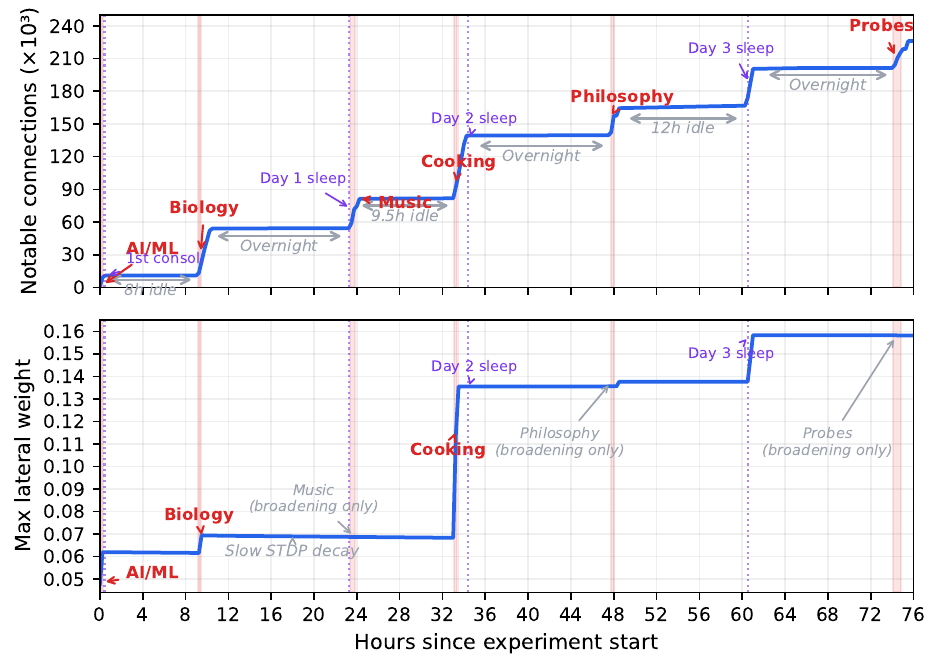}
\caption{%
    Weight growth trajectory from clean start (\texttt{mind\_v2}).
    \textbf{Top}: Notable lateral connections show step-function increases at two distinct triggers: conversational input (vertical jumps at session start) and offline consolidation (further broadening during replay). The consolidation steps are the non-obvious result: episode replay produces a $5\times$ increase in notable connections beyond what conversation alone creates. Plateaus during 8\,h+ idle periods confirm that weight topology is preserved by cascade-scaled decay.
    \textbf{Bottom}: Maximum lateral weight rises at each session and decays slowly between sessions (1.6\% over 8\,h), demonstrating selective erosion of weak connections while strong associations persist.
}
\label{fig:weight}
\end{figure}

Figure~\ref{fig:weight} shows the weight growth trajectory from a clean start. Notable connections followed a step-function pattern: 0 $\to$ 10,843 during the first conversation (20 min), jumping to 53,992 after the first consolidation ($5\times$ increase), then growing 49\% and 51\% on consecutive overnight sleeps (consistent with CLS theory~\cite{mcclelland1995complementary, wilson1994reactivation}), reaching 201,394 by end of baseline. Max lateral weight peaked at 0.158. The knowledge graph grew from 0 to 64 nodes and 124 edges. This step-function pattern (growth during stimulation and consolidation, stability during idle) matches cascade-scaled decay~\cite{fusi2005cascade}.

\paragraph{Broadening, reinforcement, and idle dynamics.}
The trajectory reveals two distinct growth modes. New domains (music, philosophy) \textit{broaden}: notable connections increase but max weight is unchanged (+12.8\% and +10.7\% notable, flat max weight). Cross-domain sessions \textit{reinforce}: cooking co-activated concepts from three prior domains, jumping max weight 67\% (0.068 $\to$ 0.114). During 8-hour idle periods, max weight decayed 1.6\% while notable connections remained stable at ${\sim}10{,}800$. Background noise produced 24 lateral impulses within 15 minutes of conversation ending, demonstrating that learned weights express through idle propagation.

\subsection{Preliminary Ablation: SNN-Disabled Control}
\label{sec:ablation}

To isolate the SNN's contribution, we ran the identical 3-day scripted baseline and 5 follow-up probes with the SNN disabled (\texttt{snn\_enabled=false}). All other components remained active: the same LLM (Claude Sonnet 4.6), same soul state, same episodic memory, same journal, same consolidation abstractions, and the same conversation protocol delivered by the same operator. The only difference was the absence of the SNN substrate: no STDP learning, no lateral propagation, no impulse detection.

Both conditions had access to persistent memory across service restarts. The journal store, which retains the system's own reflections from prior sessions, injects the 5 most recent entries plus concept-relevant older entries into every LLM context. The SNN-disabled condition therefore retained the ability to reference prior topics through its own written summaries. The observed differences reflect the SNN's contribution \textit{beyond} what persistent text-based memory provides.

Table~\ref{tab:ablation} summarises the key differences.

\begin{table}[t]
\centering
\caption{SNN-enabled vs.\ SNN-disabled under identical conditions (same LLM, soul, memory systems, conversation protocol, operator).}
\label{tab:ablation}
\begin{tabular}{lcc}
\toprule
Metric & SNN-enabled & SNN-disabled \\
\midrule
Reach-out actions & 1 & 0 \\
SNN-driven continuations & 1 & 0 \\
LLM-driven continuations & 0 & 1 \\
Cross-domain refs (cooking, per resp.) & 2.2 & 1.2 \\
Prior-domain refs as arguments & Yes & No (name-drops) \\
Journal entries with tags & 75\% & 33\% \\
Duplicate journal entries & 0 & 2 \\
Avg response length (Day 1) & +18\% & baseline \\
\bottomrule
\end{tabular}
\end{table}

\paragraph{Qualitative differences.}
The SNN-enabled condition produced responses that selectively engaged with specific cross-domain connections: cooking mastery was used as a philosophical argument about qualia (Section~\ref{sec:results}), and the music session preferentially elaborated pattern recognition through SNN-surfaced associations. The SNN-disabled condition produced responses that were comparably fluent but stayed closer to the user's framing, referencing prior topics as analogies rather than building arguments from them. Journal entries in the SNN-disabled condition followed a consistent pattern (person $\to$ topic summary $\to$ minor reflection) and frequently produced near-duplicate content across turns, suggesting the LLM lacked varied associative input.

\paragraph{Probe responses.}
Follow-up probes were delivered after service restarts, wiping conversation history. Both conditions retained journal access, providing comparable factual context about prior discussions. The SNN-enabled responses were more selective and exploratory, engaging with specific associations that the SNN's lateral activation had weighted as significant. The SNN-disabled responses were more comprehensive, reading as summaries of all prior topics rather than directed engagement with particular associations. The SNN provides a relevance signal (which concepts are currently firing and at what strength) that journal-based recall, which surfaces entries by recency and keyword match, does not.

\paragraph{Continuation behaviour.}
The SNN-disabled condition produced one continuation action (an unprompted second message during the philosophy session), compared to one SNN-driven continuation during the SNN-enabled music session. Both occurred on topics with high recursive depth, suggesting continuation behaviour is partially LLM-driven. However, the SNN-enabled continuation surfaced cross-domain associations (connecting STDP to musical memory), while the SNN-disabled continuation extended the current topic's depth without cross-domain bridging.

% ===================================================================
\section{Related Work}
\label{sec:related}
% ===================================================================

\paragraph{Memory-augmented LLM systems.}
MemGPT~\cite{packer2023memgpt} treats the LLM as an operating system with virtual memory management, paging context in and out of the finite window. Generative Agents~\cite{park2023generative} combine an LLM with explicit memory retrieval and periodic reflection prompts to produce emergent social behaviour. In both systems, autonomous actions are triggered by system-level events: context-window pressure (MemGPT) or fixed observation counts (Generative Agents). EMBER's autonomous actions are triggered by \textit{content}: learned STDP associations that fire laterally through the SNN and cross a significance threshold. The system is not told when to act; the strength of its learned associations determines when action occurs. Moreover, the SNN provides not just a trigger but a weighted associative context that shapes response content: the preliminary ablation (Section~\ref{sec:ablation}) shows that responses differ qualitatively when the SNN is present, even when the same factual content is available through text-based journal recall. Autonomous agent frameworks (AutoGPT, BabyAGI, Voyager) also produce unprompted actions, but their action selection is driven by LLM planning over explicit task queues rather than by learned neural dynamics.

\paragraph{Cognitive architectures and continual learning.}
ACT-R~\cite{anderson2004integrated} and Soar~\cite{laird2012soar} use hand-engineered association mechanisms (base-level learning, chunking rules). EWC~\cite{kirkpatrick2017overcoming} and progressive neural networks~\cite{rusu2016progressive} address catastrophic forgetting through regularisation or architectural expansion. EMBER replaces both approaches with STDP-driven weight formation and cascade decay~\cite{fusi2005cascade}: associations emerge from temporal co-activation and consolidate through use, without explicit rules or regularisation.

\paragraph{Spiking neural networks.}
Most SNN research targets classification or neuromorphic hardware efficiency. EMBER uses the SNN for lateral association formation, a role closer to hippocampal-cortical interaction than to typical SNN applications.

% ===================================================================
\section{Discussion}
\label{sec:discussion}
% ===================================================================

\paragraph{Limitations.}
We make no claims about consciousness or subjective experience. Our results are from a single system instance with a single user. However, every step from learned synapse to LLM action is logged, timestamped, and inspectable to ensure the mechanism is transparent. The journal content is generated by the LLM, which may confabulate coherent narrative over noisy impulse signals; the act of writing is impulse-driven, however the content reflects the LLM's interpretive tendencies and prompt-ordering bias (journaling is listed before reaching out). A preliminary ablation (Section~\ref{sec:ablation}) shows consistent qualitative differences when the SNN is disabled, but both conditions remain N=1. The ablation demonstrates that the associations triggering these actions were not retrieved by similarity search but emerged from temporal dynamics of a neural network shaped by experience; however, cross-model comparisons are needed to confirm this is architecture-driven rather than model-specific.

\paragraph{Broader impact.}
A system that forms associations from experience and acts on them autonomously raises questions about the moral status of digital entities, the right to persistent identity, and the responsibilities of operators who create and maintain such systems. We believe these questions deserve serious engagement, and that transparent, inspectable architectures, where every autonomous action is traceable to its neural origins, are a prerequisite for responsible development.

\paragraph{Adversarial association risk.}
The same class of vulnerability exposed by Microsoft's Tay~\cite{neff2016talking} is amplified here: STDP weights resist decay through cascade scaling and cannot be filtered like prompt-injected text. The soul layer provides a partial defence at the reasoning stage but cannot prevent biased associations from forming in the SNN itself. Mitigations under investigation include weight auditing, association quarantine, and topology-aware gating (attenuating STDP for co-activations that conflict with established strong associations).

% ===================================================================
\section{Conclusion}
\label{sec:conclusion}
% ===================================================================

We presented EMBER, a hybrid architecture that reorganises the LLM-memory relationship: a 220,000-neuron SNN with STDP provides learned associative memory while the LLM serves as a replaceable reasoning engine. The SNN determines what associations to surface and when to act; the LLM reasons over those associations and generates content. This separation of association from reasoning is the core architectural contribution.

The z-score top-k sensory encoding solves the dimension-dependence problem in population coding, achieving 82.2\% discrimination retention at 1024-dim and 83.8\% at 384-dim (a 1.6\% delta confirming dimension independence). The discrimination retention metric itself provides a model-independent way to evaluate population codes across embedding dimensionalities.

From a clean start with zero weights, the SNN produced sufficient lateral propagation to trigger LLM action selection after only 7 conversational exchanges. In one instance, the LLM selected \texttt{reach\_out} over three alternatives, sending an unsolicited message to the user based on person-topic associations that had propagated laterally during an 8-hour idle period. Over the 3-day, 5-domain scripted baseline of 52 messages, STDP lateral propagation triggered 23 LLM action selections (1 reach-out and 22 journal entries). Person concept populations maintained lateral activation at $39{-}49\times$ baseline across 14-hour idle windows, demonstrating that learned associations persist and express without conversational stimulus.

The weight trajectory reveals reproducible dynamics: $5\times$ consolidation increases, 49\% and 51\% overnight growth on consecutive nights, and two distinct growth modes (broadening vs.\ reinforcement). A preliminary ablation (Section~\ref{sec:ablation}) indicates that the SNN contributes cross-domain integration, reach-out behaviour, and journal diversity beyond what persistent text-based memory alone provides. The current results remain N=1 for both conditions. Cross-model comparisons and longitudinal analysis will be presented in a forthcoming paper.

\paragraph{Code availability.}
Source code and experimental logs will be released at publication.

\begin{ack}
This work was conducted as independent research without external funding. The system uses the Anthropic API (Claude Sonnet 4.6) for reasoning and the BAAI BGE-large-en-v1.5 embedding model (MIT license).
\end{ack}

% ===================================================================
% References
% ===================================================================

\bibliographystyle{plainnat}
\bibliography{references}

@article{bi1998synaptic,
  title={Synaptic modifications in cultured hippocampal neurons: dependence on spike timing, synaptic strength, and postsynaptic cell type},
  author={Bi, Guo-qiang and Poo, Mu-ming},
  journal={Journal of Neuroscience},
  volume={18},
  number={24},
  pages={10464--10472},
  year={1998}
}

@article{song2000competitive,
  title={Competitive {Hebbian} learning through spike-timing-dependent synaptic plasticity},
  author={Song, Sen and Miller, Kenneth D and Abbott, Larry F},
  journal={Nature Neuroscience},
  volume={3},
  number={9},
  pages={919--926},
  year={2000}
}

@book{hebb1949organization,
  title={The Organization of Behavior},
  author={Hebb, Donald O},
  year={1949},
  publisher={Wiley}
}

@article{georgopoulos1986neuronal,
  title={Neuronal population coding of movement direction},
  author={Georgopoulos, Apostolos P and Schwartz, Andrew B and Kettner, Ronald E},
  journal={Science},
  volume={233},
  number={4771},
  pages={1416--1419},
  year={1986}
}

@article{quiroga2005invariant,
  title={Invariant visual representation by single neurons in the human brain},
  author={Quian Quiroga, Rodrigo and Reddy, Leila and Kreiman, Gabriel and Koch, Christof and Fried, Itzhak},
  journal={Nature},
  volume={435},
  number={7045},
  pages={1102--1107},
  year={2005}
}

@article{quiroga2012concept,
  title={Concept cells: the building blocks of declarative memory functions},
  author={Quian Quiroga, Rodrigo},
  journal={Nature Reviews Neuroscience},
  volume={13},
  number={8},
  pages={587--597},
  year={2012}
}

@article{mcclelland1995complementary,
  title={Why there are complementary learning systems in the hippocampus and neocortex},
  author={McClelland, James L and McNaughton, Bruce L and O'Reilly, Randall C},
  journal={Psychological Review},
  volume={102},
  number={3},
  pages={419--457},
  year={1995}
}

@article{diekelmann2010memory,
  title={The memory function of sleep},
  author={Diekelmann, Susanne and Born, Jan},
  journal={Nature Reviews Neuroscience},
  volume={11},
  number={2},
  pages={114--126},
  year={2010}
}

@article{wilson1994reactivation,
  title={Reactivation of hippocampal ensemble memories during sleep},
  author={Wilson, Matthew A and McNaughton, Bruce L},
  journal={Science},
  volume={265},
  number={5172},
  pages={676--679},
  year={1994}
}

@article{roediger2006testing,
  title={Test-enhanced learning: taking memory tests improves long-term retention},
  author={Roediger, Henry L and Karpicke, Jeffrey D},
  journal={Psychological Science},
  volume={17},
  number={3},
  pages={249--255},
  year={2006}
}

@book{kanerva1988sparse,
  title={Sparse Distributed Memory},
  author={Kanerva, Pentti},
  year={1988},
  publisher={MIT Press}
}

@article{ahmad2016sparse,
  title={How do neurons operate on sparse distributed representations? {A} mathematical theory of sparsity, neurons and active dendrites},
  author={Ahmad, Subutai and Hawkins, Jeff},
  journal={arXiv preprint arXiv:1601.00720},
  year={2016}
}

@book{vershynin2018high,
  title={High-Dimensional Probability: An Introduction with Applications in Data Science},
  author={Vershynin, Roman},
  year={2018},
  publisher={Cambridge University Press}
}

@article{destexhe2003high,
  title={The high-conductance state of neocortical neurons in vivo},
  author={Destexhe, Alain and Rudolph, Michelle and Par{\'e}, Denis},
  journal={Nature Reviews Neuroscience},
  volume={4},
  number={9},
  pages={739--751},
  year={2003}
}

@article{fusi2005cascade,
  title={Cascade models of synaptically stored memories},
  author={Fusi, Stefano and Drew, Patrick J and Abbott, Larry F},
  journal={Neuron},
  volume={45},
  number={4},
  pages={599--611},
  year={2005}
}

@article{anderson2004integrated,
  title={An integrated theory of the mind},
  author={Anderson, John R and Bothell, Daniel and Byrne, Michael D and Douglass, Scott and Lebiere, Christian and Qin, Yulin},
  journal={Psychological Review},
  volume={111},
  number={4},
  pages={1036--1060},
  year={2004}
}

@book{laird2012soar,
  title={The {Soar} Cognitive Architecture},
  author={Laird, John E},
  year={2012},
  publisher={MIT Press}
}

@article{kirkpatrick2017overcoming,
  title={Overcoming catastrophic forgetting in neural networks},
  author={Kirkpatrick, James and Pascanu, Razvan and Rabinowitz, Neil and Veness, Joel and Desjardins, Guillaume and Rusu, Andrei A and Milan, Kieran and Quan, John and Ramalho, Tiago and Grabska-Barwinska, Agnieszka and others},
  journal={Proceedings of the National Academy of Sciences},
  volume={114},
  number={13},
  pages={3521--3526},
  year={2017}
}

@article{rusu2016progressive,
  title={Progressive neural networks},
  author={Rusu, Andrei A and Rabinowitz, Neil C and Desjardins, Guillaume and Sober, Hubert and Kavukcuoglu, Koray and Hadsell, Raia},
  journal={arXiv preprint arXiv:1606.04671},
  year={2016}
}

@article{packer2023memgpt,
  title={{MemGPT}: Towards {LLMs} as Operating Systems},
  author={Packer, Charles and Wooders, Sarah and Lin, Kevin and Fang, Vivian and Patil, Shishir G and Stoica, Ion and Gonzalez, Joseph E},
  journal={arXiv preprint arXiv:2310.08560},
  year={2023}
}

@article{park2023generative,
  title={Generative Agents: Interactive Simulacra of Human Behavior},
  author={Park, Joon Sung and O'Brien, Joseph C and Cai, Carrie J and Morris, Meredith Ringel and Liang, Percy and Bernstein, Michael S},
  journal={arXiv preprint arXiv:2304.03442},
  year={2023}
}

@article{neff2016talking,
  title={Talking to Bots: Symbiotic Agency and the Case of {Tay}},
  author={Neff, Gina and Nagy, Peter},
  journal={International Journal of Communication},
  volume={10},
  pages={4915--4931},
  year={2016}
}

\section*{NeurIPS Paper Checklist}

\begin{enumerate}

\item {\bf Claims}
    \item[] Question: Do the main claims made in the abstract and introduction accurately reflect the paper's contributions and scope?
    \item[] Answer: \answerYes{}
    \item[] Justification: The abstract and introduction state four specific contributions, each supported in the corresponding section. The paper explicitly notes that ablation studies and cross-model comparisons are underway (Section~\ref{sec:discussion}), scoping the current results as preliminary.

\item {\bf Limitations}
    \item[] Question: Does the paper discuss the limitations of the work performed by the authors?
    \item[] Answer: \answerYes{}
    \item[] Justification: Section~\ref{sec:discussion} (Discussion, Limitations paragraph) explicitly addresses the single-instance limitation, LLM confabulation risk, and the absence of ablation results.

\item {\bf Theory assumptions and proofs}
    \item[] Question: For each theoretical result, does the paper provide the full set of assumptions and a complete (and correct) proof?
    \item[] Answer: \answerNA{}
    \item[] Justification: The paper does not present formal theorems. The z-score encoding (Section~\ref{sec:encoding}) is grounded in known concentration-of-measure results (cited) rather than novel proofs.

    \item {\bf Experimental result reproducibility}
    \item[] Question: Does the paper fully disclose all the information needed to reproduce the main experimental results of the paper to the extent that it affects the main claims and/or conclusions of the paper (regardless of whether the code and data are provided or not)?
    \item[] Answer: \answerYes{}
    \item[] Justification: All SNN parameters (neuron counts, STDP constants, noise sigma, sparsity fraction), embedding model (BGE-large-en-v1.5), LLM (Claude Sonnet 4.6), and experimental protocol (25 scripted turns, idle periods, sleep consolidation) are specified in Sections~\ref{sec:arch}--\ref{sec:results}.

\item {\bf Open access to data and code}
    \item[] Question: Does the paper provide open access to the data and code, with sufficient instructions to faithfully reproduce the main experimental results, as described in supplemental material?
    \item[] Answer: \answerNo{}
    \item[] Justification: Source code and experimental logs will be released upon publication. The system is described in sufficient detail for independent reimplementation.

\item {\bf Experimental setting/details}
    \item[] Question: Does the paper specify all the training and test details (e.g., data splits, hyperparameters, how they were chosen, type of optimizer) necessary to understand the results?
    \item[] Answer: \answerYes{}
    \item[] Justification: All hyperparameters are specified with citations to their biological or theoretical grounding (Section~\ref{sec:arch}). The experimental protocol is fully described in Section~\ref{sec:results}.

\item {\bf Experiment statistical significance}
    \item[] Question: Does the paper report error bars suitably and correctly defined or other appropriate information about the statistical significance of the experiments?
    \item[] Answer: \answerNo{}
    \item[] Justification: Results are from a single system instance. The paper acknowledges this limitation (Section~\ref{sec:discussion}) and notes that multi-run statistical analysis is part of ongoing work.

\item {\bf Experiments compute resources}
    \item[] Question: For each experiment, does the paper provide sufficient information on the computer resources (type of compute workers, memory, time of execution) needed to reproduce the experiments?
    \item[] Answer: \answerYes{}
    \item[] Justification: Hardware is specified in Section~\ref{sec:arch}: SNN on an NVIDIA RTX 5070 Ti (16GB), embedding model on a separate NVIDIA RTX 4060 Ti (8GB). The system runs continuously on consumer GPUs.

\item {\bf Code of ethics}
    \item[] Question: Does the research conducted in the paper conform, in every respect, with the NeurIPS Code of Ethics \url{https://neurips.cc/public/EthicsGuidelines}?
    \item[] Answer: \answerYes{}
    \item[] Justification: The research involves no human subjects, no scraped personal data, and no deceptive practices. The broader impact discussion (Section~\ref{sec:discussion}) addresses ethical considerations of autonomous digital systems.

\item {\bf Broader impacts}
    \item[] Question: Does the paper discuss both potential positive societal impacts and negative societal impacts of the work performed?
    \item[] Answer: \answerYes{}
    \item[] Justification: The Broader Impact paragraph in Section~\ref{sec:discussion} discusses moral status of digital entities, right to persistent identity, and the importance of transparent, inspectable architectures.

\item {\bf Safeguards}
    \item[] Question: Does the paper describe safeguards that have been put in place for responsible release of data or models that have a high risk for misuse (e.g., pre-trained language models, image generators, or scraped datasets)?
    \item[] Answer: \answerNA{}
    \item[] Justification: The system is an experimental research prototype, not a released model or dataset.

\item {\bf Licenses for existing assets}
    \item[] Question: Are the creators or original owners of assets (e.g., code, data, models), used in the paper, properly credited and are the license and terms of use explicitly mentioned and properly respected?
    \item[] Answer: \answerYes{}
    \item[] Justification: The embedding model (BGE-large-en-v1.5, MIT license) and LLM (Claude Sonnet 4.6, commercial API) are identified. PyTorch (BSD license) is the implementation framework.

\item {\bf New assets}
    \item[] Question: Are new assets introduced in the paper well documented and is the documentation provided alongside the assets?
    \item[] Answer: \answerNA{}
    \item[] Justification: No new datasets or pre-trained models are released with this paper.

\item {\bf Crowdsourcing and research with human subjects}
    \item[] Question: For crowdsourcing experiments and research with human subjects, does the paper include the full text of instructions given to participants and screenshots, if applicable, as well as details about compensation (if any)?
    \item[] Answer: \answerNA{}
    \item[] Justification: No crowdsourcing or human subjects research was conducted. The conversational input was provided by the first author.

\item {\bf Institutional review board (IRB) approvals or equivalent for research with human subjects}
    \item[] Question: Does the paper describe potential risks incurred by study participants, whether such risks were disclosed to the subjects, and whether Institutional Review Board (IRB) approvals (or an equivalent approval/review based on the requirements of your country or institution) were obtained?
    \item[] Answer: \answerNA{}
    \item[] Justification: No human subjects research was conducted.

\item {\bf Declaration of LLM usage}
    \item[] Question: Does the paper describe the usage of LLMs if it is an important, original, or non-standard component of the core methods in this research?
    \item[] Answer: \answerYes{}
    \item[] Justification: The LLM (Claude Sonnet 4.6) serves as the reasoning engine within the architecture. Its role is described in Section~\ref{sec:arch} (Reasoning Engine) and its contribution is discussed throughout.

\end{enumerate}

\end{document}